\documentclass[10pt,twocolumn,letterpaper]{article}

\usepackage{cvpr}              

\usepackage{graphicx}
\usepackage{amsmath}
\usepackage{amssymb}
\usepackage{booktabs}
\usepackage{float}

\usepackage[pagebackref,breaklinks,colorlinks]{hyperref}

\usepackage[capitalize]{cleveref}
\crefname{section}{Sec.}{Secs.}
\Crefname{section}{Section}{Sections}
\Crefname{table}{Table}{Tables}
\crefname{table}{Tab.}{Tabs.}

\def\cvprPaperID{*****} 
\def\confName{CVPR}
\def\confYear{2022}

\begin{document}

\title{Out of Distribution Performance of State of Art Vision Model}

\author{Salman Rahman\\
NYU\\
{\tt\small salman@nyu.edu}
\and
Wonkwon Lee\\
NYU\\
{\tt\small wl2733@nyu.edu}
}
\maketitle

\begin{abstract}
   The vision transformer (ViT) has advanced to the cutting edge in the visual recognition task. Transformers are more robust than CNN, according to the latest research. ViT's self-attention mechanism, according to the claim, makes it more robust than CNN. Even with this, we discover that these conclusions are based on unfair experimental conditions and just comparing a few models, which did not allow us to depict the entire scenario of robustness performance. In this study, we investigate the performance of 58 state-of-the-art computer vision models in a unified training setup based not only on attention and convolution mechanisms but also on neural networks based on a combination of convolution and attention mechanisms, sequence-based model, complementary search, and network-based method. Our research demonstrates that robustness depends on the training setup and model types, and performance varies based on out-of-distribution type. Our research will aid the community in better understanding and benchmarking the robustness of computer vision models. The code and models are publicly available.\footnote{\url{https://github.com/salman-lui/vision_course_project}}
\end{abstract}

\section{Introduction and Motivating Works}
\label{sec:intro}
The shift in data distribution is a challenging machine-learning problem, especially in computer vision. Distribution shifts are caused by poor data quality, out-of-distribution, and adversarial problems. Transformer is a powerful visual recognition tool. Recent research also claims that Transformers are far more robust than Convolutional Neural Networks, as well as having competitive performance on several visual benchmarks (CNNs) \cite{paul2022vision, mahmood2021robustness}.


Natural language processing dramatically advanced thanks to the Transformers invention by Vaswani et al. in 2017 \cite{vaswani2017attention}. With the addition of the self-attention module, the Transformer can now accurately grasp the non-local interactions between all of the input sequence parts and achieve cutting-edge performance on various NLP tasks \cite{dai2019transformer, brown2020language,devlin2018bert, yang2019xlnet}.

The Transformer's success in NLP is beginning to be seen in computer vision \cite{dosovitskiy2020image, khan2022transformers}. The ground-breaking work, Vision Transformer (ViT), shows that the pure Transformer architectures can perform astounding on many visual benchmarks, mainly when massive datasets (like JFT-300M) are available for pre-training \cite{sun2017revisiting}. The tokenization module of Transformers is enhanced, multi-resolution feature maps are built on Transformers, parameter-efficient Transformers are designed for scalability, and other improvements are made after carefully curating the training pipeline and distillation method \cite{touvron2021going, xue2022go}. Our goal in this work is to present a fair and fully comprehend of the out-of-distribution performance of almost all the states of the computer vision model, each of which is based on a unique recipe.

Paul et al. compare the performance of ViT models with the state-of-the-art convolutional neural networks (CNNs), Big-Transfer, in a thorough performance comparison in order to investigate the robustness of the Vision Transformer \cite{paul2022vision}. ViT is a robust learner, according to the analysis, which the author credits to its use of large-scale pre-training and attention modules, its capacity to recognize images that have been randomly masked, its low sensitivity to perturbations in the Fourier spectrum domain, and its ability to exhibit a wider energy distribution and a smoother loss landscape in the face of adversarial input perturbations. Mahmood et al. find that vision transformers are robust to adversarial examples \cite{mahmood2021robustness}. Naseer et al. investigate the intriguing properties of vision transformers and find that ViTs are highly robust to severe occlusions, perturbations, and domain shifts \cite{naseer2021intriguing}. Bhojanapalli et al. discover that ViT models, when pre-trained with enough data, are at least as robust to various perturbations as their ResNet counterparts \cite{bhojanapalli2021understanding}. 

Transformers are supposedly more robust than CNNs, according to the literature currently in circulation, although different sets of the model require a different training setup for comparison. Additionally, to our knowledge, there is no literature that examines the performance of each SOTA vision model's robustness. We compare the robustness performance of vision models built upon various mechanisms, such as convolution, attention, multi-layer perceptron, MobileNet, and sequence, in this work by trying to adhere to a unified training setup for not only CNN and the ViT model but also all other existing vision models.

\section{Methods}
\label{sec:method}

\subsection{Design and Settings}
We aim to provide fair and in-depth comparisons Transformers, CNNs, and other combinations on robustness evaluations. Our models are tested on a variety of benchmark robustness datasets, including ImageNet-A, ImageNet-R, ImageNet-O, ImageNet-Sketch, and Stylized ImageNet. Recent research on robustness comparison between Transformers and CNNs is largely conducted in unfair experimental settings, where models are not compared at the same scale of parameters. For example, a comparison between a small CNN ResNet-50 with 25 million parameters and a large Transformer ViT-B with 86 million parameters is not fair evaluation. Also district training frameworks are applied to distinct models, in terms of training datasets, number of epochs, and augmentation strategies. In our research, we evaluate small baseline models of each of the recipes in a fair experimental environment of out-of-distribution benchmark sets. 

\subsubsection{Convolution Based Model Training}
Convolutional neural network 
Among the 29 state-of-the-art convolutional neural network models, ResNet-50, DenseNet121, EfficientNet-B, and TinyNet-A are implemented following the standard recipes. All CNNs are trained on ImageNet using Adam optimizer and an initial learning rate of 0.01. For the rest of the models, pre-trained models are used to evaluate the robustness performance due to the challenge in  training.

\subsubsection{Attention Based Model Training}
By effectively transferring Transformers from natural language processing to computer vision, ViT outperforms CNNs on a number of visual benchmarks. In this experiment, DeiT-S, BEiT-B, Visformer-S, and PiT-B are implemented as the default Transformer architecture. All transformers are trained on ImageNet using Adam optimizer and an initial learning rate of 0.01. 

The same training framework used on CNN models is also used on our Transformer models (e.g., 100 training epochs, the same optimizer and learning rate, etc). Pre-trained weights are used for the rest of the Transformer models due to the challenge of training.

\subsubsection{Combined Attention and Convolution Based Training }
ConViT is a state-of-the-art ViT enhanced with soft convolutional inductive biases. The model is constructed based on the DeiT-B, an open-sourced hyperparameter-optimized version of the ViT. The $3 \times 3$ convolutional filters are applied in the form of 6 attention heads as in \cite{touvron2021training}. For other combined models like ConvNext, EdgeNeXt, and LeViT, baseline pre-trained models are used for faster evaluation.

\subsubsection{Model (not Convolution or Attention) Training}
For models with a combination of other models (not convolution-based or attention-based), such as MLP-Mixer, MobileNet, RegNet, and Sequencer, pre-trained weights are used for convenience \cite{rw2019timm}. 

\subsection{Out of Distribution Robustness}

Several benchmarks using out-of-distribution samples have recently been presented to assess how deep neural networks perform when tested outside the box. Two such benchmarks are present in our robustness evaluation suite in particular: \begin{itemize}
    \item \textbf{ImageNet - A:} ImageNet-A \cite{hendrycks2021natural} contains real-world images gathered from scenes with complex recognizing tasks like occlusion and fog. Such images can lead to misclassification for reasons such as multiple objects associated with to a single discrete label. In table \ref{tab:cnn-1} and \ref{tab:vit-1}, we report the top-1 accuracy of various models on the ImageNet-A dataset. MViT v2 achieves the highest top-1 accuracy score on the dataset, while MobileViT only reports the lowest score. Transformer models show superior performance than CNN and other models, and thus self-attention is considered a key to solving these problems.
    
    \item \textbf{ImageNet - R:} ImageNet-R(endition) \cite{hendrycks2021many} contains 30000 images containing various renditions (e.g., art, cartoons, graphics, paintings, etc) of ImageNet 1000 object classes. Such representations are naturally occurring and differ from ImageNet images in terms of textures and local image statistics, allowing us measures the synthetic robustness of the vision network by semantic shifts under various domains. In our experiment, the attention-based models report a superior performance over the convolution-based and other models. Among attention-based models, DeiT3 achieves the highest accuracy of 53.63\%, and EfficientNet V2 achieves the highest of 52.97\% among convolution-based models.
    
    \item \textbf{ImageNet - O:}  ImageNet-O \cite{hendrycks2021natural} comprises images from 200 different classes that a model did not observe during training and are regarded as anomalies. A robust model is expected to achieve low confidence scores on such images. The dataset is adversarially filtered samples for ImageNet out-of-distribution detectors and contains the 200-class subset of ImageNet's 1000 classes. For the evaluation strategy, the \textit{area under the precision-recall curve} (AUPR) is used as in the paper \cite{hendrycks2021natural}. MobileViT and BEiT report the highest out-of-distribution scores among all the models.  
    

    \item \textbf{ImageNet - Sketch:} ImageNet-Sketch \cite{wang2019learning} contains 50000 black and white sketch images, 50 images for each of the 1000 ImageNet classes. It is a unique ImageNet-scale out-of-domain evaluation dataset for classification and the images are gathered independently from the original validation set. Such out-of-domain black-and-white sketch images can measure a model's ability to extrapolate out of domain. 
    
    \item \textbf{Stylized ImageNet:} Stylized ImageNet \cite{geirhos2018} is a stylized version of ImageNet generated by AdaIN style transfer such as greyscale, silhouette, edges, texture, and etc. While global object shapes remain intact after stylization, local textures are severely distorted. This allows Stylized-ImageNet to induce CNNs to learn about shapes and less about local textures.
\end{itemize}

\section{Robustness Experiments and Results}
\label{sec:experiments}
We conduct a thorough robustness analysis (on six out of distribution data) of 58 cutting-edge computer vision models based on different mechanisms, including convolution \cite{he2016deep}, attention \cite{dosovitskiy2020image}, attention \& convolution combined \cite{d2021convit}, multi-layer perceptrons\cite{tolstikhin2021mlp}, complementary search methods (MobileNet)\cite{howard2019searching}, networks (RegNet)\cite{radosavovic2020designing}, and sequence\cite{tatsunami2022sequencer}.

\subsection{Convolution Mechanism}
Based on a comprehensive convolution process inspired by LeNet \cite{lecun1998gradient} and AlexNet \cite{krizhevsky2017imagenet}, we choose 29 state-of-the-art computer vision models along with their accuracy listed in Table \ref{tab:cnn-1} and \ref{tab:cnn-2} alphabetically. 


ConvNeXt is a pure ConvNet model that can compete favorably across several computer vision benchmarks with state-of-the-art hierarchical vision Transformers while retaining the simplicity and efficiency of standard ConvNets \cite{liu2022convnet}. It achieves the highest ImageNet top-1 accuracy with 87.8\%. It also reports 35.79\% accuracy on ImageNet-A, 19.24\% and 38.22\% on Stylized ImageNet and Sketch. ConvNeXt demonstrates a better out-of-distribution classification than Transformers and others. 
Using a cross-stage partial network, Wang et al. CSPNet solution reduced computation by 20\% without sacrificing accuracy while addressing the issue of high inference computations from the perspective of network design \cite{wang2020cspnet}. While CSPNet's ImageNet performance is 77.9\%, testing on ImageNet-A and ImageNet-O reveals a dramatic decline in performance. Yet, CSPNet reports fairly high scores in ImageNet-R, Stylized ImageNet, and ImageNet Sketch among CNNs, even comparable to Transformers. Another kind of neural net, DenseNet solves the vanishing-gradient issue, strengthens feature propagation, promotes feature reuse, and significantly lowers the number of parameters by connecting each layer to every other layer in a feed-forward fashion \cite{huang2017densely}. DenseNet's ImageNet performance is 77.36\%, yet it decreases drastically when evaluated on out-of-distribution datasets.

\begin{table}
\centering
\caption{Performance of convolution-based model on ImageNet, ImageNet-A, and ImageNet-O}
\label{tab:cnn-1}
\begin{tabular}{cccc} 
\hline
\textbf{Architecture} & \textbf{ImageNet} & \begin{tabular}[c]{@{}c@{}}\textbf{ImageNet}\\\textbf{ -A}\end{tabular} & \begin{tabular}[c]{@{}c@{}}\textbf{ImageNet}\\\textbf{ -O}\end{tabular}  \\ 
\hline
ConvNext    & 87.8   & 35.79   & 2.18  \\
CSPNet             & 77.90       & 6.71         & 2.19                    \\
DenseNet           & 77.36        & 4.97         & 2.17                    \\
DLA34              & 79.44       & 3.21         & 2.22                    \\
DPN68              & 80.16       & 3.63         & 2.23                    \\
EfficientNetV2     & 87.3       & 22.31        & 2.21                    \\
EfficientNet       & 77.10       & 7.07         & 2.18                    \\
MIXNet             & 78.90       & 8.43         & 2.17                    \\
MnasNet            & 75.2       & 2.88         & 2.16                    \\
FBNet              & 74.10       & 4.09         & 2.16                    \\
SPNASNet           & 74.96       & 3.57         & 2.18                    \\
TinyNet            & 59.90       & 7.89         & 2.18                    \\
GhostNet           & 75.70       & 3.24         & 2.18                    \\
HRNet              & 77.30        & 5.51         & 2.23                    \\
InceptionResNetV2  & 81.30       & 5.72         & 2.25                    \\
InceptionV3        & 78.80       & 5.65         & 2.29                    \\
NASNet-A           & 82.70       & 3.49         & 2.97                    \\
NFResNet           & 75.90       & 6.77         & 2.17                    \\
PNASNet            & 74.20       & 3.39         & 2.34                    \\
Res2Net            & 79.19      & 5.05         & 2.11                    \\
ResNeSt            & 83.90       & 5.47         & 2.2                     \\
ResNet             & 79.29       & 1.88         & 2.23                    \\
ReXNet             & 81.63       & 8.8          & 2.17                    \\
SelecSLS           & 78.40       & 4.53         & 2.16                    \\
SKNet           & 80.15       & 2.75         & 2.17                    \\
TResNet            & 80.80       & 17.17        & 2.2                     \\
VGG19              & 76.30       & 2.11         & 2.37                    \\
VoVNet             & 79.31       & 6.59         & 2.18                    \\
Xception           & 79.88       & 3.53         & 2.28                    \\
\hline
\end{tabular}
\end{table}

Deep Layer Aggregation (DLA) by Yu et al. iteratively and hierarchically combines the feature hierarchy to create networks with more accuracy and fewer parameters \cite{yu2018deep}. While DLA achieves 79.44\% accuracy in ImageNet, it reports one of the lowest scores in ImageNet-A, O, and Stylized. DLA also achieves average scores in ImageNet-R and Sketch. The state-of-the-art Residual Network (ResNet) and Densely Convolutional Network (DenseNet) feature reuse and new feature exploration systems are combined into one framework by the Dual Path Network (DPN). It employs a small model, incurs a little computational cost, and uses less GPU memory to achieve 80.16\% accuracy on ImageNet data \cite{chen2017dual}. The model under-performs with ImageNet-A and O but achieves high accuracy of 11.04\%  on Stylized-ImageNet. A novel convolutional network architecture called EfficientNetV2 has improved parameter efficiency and faster training time. These models are created by combining scaling with training-aware neural architecture search to enhance training efficiency and speed jointly \cite{tan2021efficientnetv2}. Another member of the EfficientNet family uses a simple yet very efficient compound coefficient to scale all depth, breadth, and resolution dimensions uniformly \cite{tan2019efficientnet}. 
EfficientNetV2 achieves 87.3\% on ImageNet and reports the highest score on ImageNet-A, R, Stylized, and Sketch among CNNs, with 22.31\%, 52.97\%, 23.3\%, and 40.33\% accuracy respectively. Further, its performance on ImageNet-R and Stylized is the second highest, and the performance on ImageNet-Sketch is the highest among all the tested models, showing that CNNs can be more robust than Transformers.
EfficientNet achieves 77.1\% on ImageNet, but it decreases dramatically on robustness datasets.

In a study of the effects of various kernel sizes for depthwise convolution, MIXNet finds that the single kernel size constraint is a problem for conventional depthwise convolution \cite{tan2019efficientnet}. This was the driving force for the author's suggestion of a mixed depthwise convolution, which incorporates several kernel sizes into a single convolution.
MIXNet achieves the third-highest ImageNet-A score and fairly high performance on other datasets as well. 
Model latency is explicitly incorporated into the primary purpose of Tan et al. Mobile Neural Architecture Search (MnasNet) proposal so that the search can find a model that successfully balances accuracy and latency \cite{tan2019mnasnet}. Wu et al. suggests a differentiable neural architecture search (DNAS) framework that employs gradient-based techniques to optimize ConvNet designs rather than defining and training each architecture independently as in earlier approaches. A family of models called FBNets (Facebook-Berkeley-Nets), which DNAS identified, outperforms the most advanced models created manually and automatically \cite{wu2019fbnet}. In less than 4 hours, Single-Path NAS (SPNASNet), a novel differentiable Neural architecture search approach, can create hardware-efficient ConvNets and reach top-1 accuracy of 74.96\% on ImageNet \cite{stamoulis2019single}. All MnasNet, FBNet, and SPNASNet scores a high prediction accuracy on ImageNet, yet their performance all decreased on out-of-distribution datasets.

\begin{table}[]
\caption{Performance of convolution-based model on ImageNet-R, Stylized-ImageNet, and ImageNet-Sketch}
\label{tab:cnn-2}
\begin{tabular}{cccc}
\hline
\textbf{Architecture} & \textbf{\begin{tabular}[c]{@{}c@{}}ImageNet\\ -R\end{tabular}} & \textbf{\begin{tabular}[c]{@{}c@{}}Stylized\\ -ImageNet\end{tabular}} & \textbf{\begin{tabular}[c]{@{}l@{}}ImageNet\\ -Sketch\end{tabular}} \\ \hline
ConvNext          & 51.71  & 19.24           & 38.22     \\
CSPNet            & 43.06  & 10.58           & 31.25       \\
DenseNet          & 39.19  & 8.13            & 26.83       \\
DLA34             & 35.63  & 7.61            & 23.68       \\
DPN68             & 39.31  & 11.04           & 26.18       \\
EfficientNetV2    & 52.97  & 23.3            & 40.33       \\
EfficientNet      & 37.05  & 11.45           & 25.07       \\
MIXNet            & 38.64  & 11.13           & 25.64       \\
MnasNet           & 34.35  & 7.63            & 21.67       \\
FBNet             & 34.25  & 7.66            & 21.49       \\
SPNASNet          & 33.49  & 7.39            & 20.87       \\
TinyNet           & 38.57  & 12.04           & 26.54       \\
GhostNet          & 33.19  & 9.94            & 21.63       \\
HRNet             & 38.27  & 9.05            & 25.98       \\
InceptionResNetV2 & 40.68  & 13.27           & 27.99       \\
InceptionV3       & 38.05  & 11.8            & 26.78       \\
NASNet-A          & 30.11  & 11.52           & 8.72        \\
NFResNet          & 43.41  & 11.19           & 31.48       \\
PNASNet           & 31.21  & 9.07            & 16.03       \\
Res2Net           & 38.72  & 8.85            & 26.47       \\
ResNeSt           & 37.77  & 7.39            & 25.28       \\
ResNet            & 28.93  & 3.59            & 16.84       \\
ReXNet            & 37.5   & 11.04           & 26.94       \\
SelecSLS          & 36.67  & 6.75            & 23.98       \\
SKNet          & 36.94  & 9.25            & 24.88       \\
TResNet           & 39.23  & 12.56           & 21.03       \\
VGG19             & 28.58  & 3.03            & 17.94       \\
VoVNet            & 35.05  & 8.5             & 23.84       \\
Xception          & 33.52  & 9.33            & 18.26       \\ \hline
\end{tabular}
\end{table}

With the FLOPs limitation, TinyNet achieves 59.9\% top-1 accuracy on ImageNet by breaking down neural architectures into a series of smaller models generated from EfficientNet-B0 \cite{han2020model}. Interestingly, TinyNet achieves one of the highest scores on ImageNet-A, R and Stylized, with 7.89\%, 38.57\%, and 12.04\% respectively. Recent deep learning networks have high computational costs. GhostNet suggests a unique Ghost module to produce more feature maps from cheap operations based on a set of intrinsic feature maps \cite{han2020ghostnet}. By connecting high-to-low-resolution convolutions in parallel and continuously performing fusions across parallel convolutions, the high-resolution network (HRNet) designed for human pose estimation preserves high-resolution representations throughout the entire process \cite{sun2019deep, sun2019high}. Both GhostNet and HRNet achieves similarly high accuracy on ImageNet, yet HRNet slightly outperforms GhostNet on all of the datasets except Stylized ImageNet.

The Inception architecture and residual connections are combined in InceptionResNetV2, which dramatically enhances recognition performance and achieves 81.3\% top-1 accuracy in ImageNet \cite{szegedy2017inception}. With the help of aggressive regularization and properly factorized convolutions, InceptionV3 obtained 78.8\% top-1 accuracy in ImageNet \cite{szegedy2016rethinking}.
Both models demonstrate high performance among CNNs, especially InceptionResNetV2 achieves the second highest top-1 accuracy on Imagenet-R (40.68\%) and Stylized (13.27\%), comparable to Transformers.

The idea of NASNet-A is to find an architectural building block on a small dataset and then transfer it to a larger dataset \cite{zoph2018learning}. A brand-new technique called Progressive Neural Architecture Search (PNASNet) is more effective than current state-of-the-art techniques based on reinforcement learning, and evolutionary algorithms for learning the structure of convolutional neural networks (CNNs) \cite{liu2018progressive}. NASNet achieves one of the highest scores of ImageNet with 82.7\% and also reports the highest AUPR of 2.97  on ImageNet-O among convolution-based models and is the third highest in all of the tested models on ImageNet-O. Although PNASNet achieves 75.9\% on ImageNet, it outperforms NASNet on ImageNet-R and Sketch.

The milestone architecture ResNet has a number of variations, and amongst NFResNet, Res2Net, ResNeSt, ResNet, and ReXNet are evaluated. To build highly effective ResNets without activation normalization layers, NFResNet presents a straightforward collection of analysis tools to quantify signal propagation on the forward pass \cite{brock2021characterizing}. Gao et al. suggest Res2Net as a unique CNN building block by creating hierarchical residual-like connections within a single residual block \cite{gao2019res2net}. ResNeSt is a modularized architecture that uses channel-wise attention to focus on various network branches to take advantage of their success in capturing cross-feature interactions and learning various representations \cite{zhang2022nested}. The state-of-the-art computer vision model, ResNet, uses a residual learning framework to make it easier to train far deeper networks than previously employed \cite{he2016deep}. RexNet is a simple yet efficient channel setup that achieves 81.63\% top-1 accuracy on ImageNet when parameterized by the layer index \cite{han2021rethinking}. All of the models achieve more than 75\% top-1 accuracy on ImageNet, though they did not perform well on ImageNet-A and O. On ImageNet-R, Stylized, and Sketch datasets, all of them except ResNet achieve high scores. Therefore, ResNet is the least robust model among its variations. Also NFResNet achieves the highest accuracy on ImageNet-R and Sketch with 43.41\% and 31.48\%.

SelecSLS improves information flow by utilizing new selective long and short-range skip connections, enabling a significantly faster network without sacrificing accuracy \cite{mehta2020xnect}. It achieves 78.4\% top-1 accuracy on ImageNet. 
A dynamic selection technique in CNN's called selective kernel networks (SKNet) enables each neuron to adaptively change the size of its receptive field based on various scales of input data. Selective Kernel (SK) units are a type of building block that fuse several branches with various kernel sizes utilizing softmax attention directed by the information in these branches \cite{li2019selective}. SKNet achieves 80.15\% accuracy on ImageNet. Yet, both SelecSLS and SKNet decrease dramatically on ImageNet -A, O, R, and others.

TResNet is a high-performance GPU-dedicated architecture that outperforms earlier ConvNets in terms of accuracy and efficiency \cite{ridnik2021tresnet}. TResNet achieves fairly high accuracy of 80.8\% on ImageNet, and it reports 17.17\% accuracy on ImageNet-A and 39.23\% on ImageNet-R, which are one of the highest scores among CNNs.

VGGNet is a deep convolutional neural network that achieves 76.3\% top-1 accuracy in ImageNet \cite{simonyan2014very}. However, VGGNet's performance on out-of-distribution robustness is below expectation, reporting the lowest Stylized ImageNet accuracy.

CenterMask, a straightforward yet effective anchor-free instance segmentation method for VoVNet, combines two successful strategies: effective Squeeze-Excitation (eSE), which addresses the issue of channel information loss associated with original Squeeze-Excitation, and residual connection, which addresses the optimization issue of larger VoVNet \cite{lee2020centermask}. In convolutional neural networks, Inception modules are interpreted by Xception as a transitional stage between the depthwise separable convolution operation, and ordinary convolution \cite{chollet2017xception}. While VoVNet and Xception achieve more than 79\% top-1 accuracy on ImageNet, their metric scores decrease on robust image classification.

\subsection{Attention Mechanism}

A self-supervised vision representation methodology called Bidirectional Encoder representation from Image Transformers (BEiT) tokenizes the original image into visual tokens, randomly masks some image patches, and decodes them into the main Transformer \cite{bao2021beit}. Although BEiT achieves 82.9\% accuracy  on ImageNet, it's accuracy on ImageNet-A, R, Stylized, and Sketch are one of the lowest, even lower than convolution-based models. Yet, it reports AUPR of 3.46 on ImageNet-O dataset, which is the second highest among all the models.

\begin{table}
\centering
\caption{Performance of attention-based vision model on ImageNet, ImageNet-A, and ImageNet-O}
\label{tab:vit-1}
\begin{tabular}{cccc} 
\hline
\textbf{Architecture} & \textbf{ImageNet} & \textbf{ImageNet-A} & \textbf{ImageNet-O}  \\ 
\hline
BEiT                  & 82.9           & 3.92       & 3.46       \\
CaiT                  & 86.5            & 34.4       & 2.2        \\
CoaT                  & 81.9           & 16.68      & 2.2        \\
CrossViT              & 81.5           & 31.8       & 2.2        \\
DeiT                  & 81.8           & 27.85      & 2.21       \\
DeiT3                 & 83.1           & 40.12      & 2.22       \\
EfficientFormer       & 79.2          & 13.93      & 2.24       \\
GCViT                 & 83.4           & 42.71      & 2.2        \\
MaxViT                & 86.5           & 23.32      & 2.2        \\
MobileViT             & 78.4           & 0.33       & 3.92       \\
MViTv2                & 88.8           & 43.09      & 2.24       \\
NesT                  & 81.5           & 34.15      & 2.22       \\
PiT                   & 82.0           & 33.89      & 2.2        \\
PVT V2                & 82.0           & 4.05       & 2.18       \\
Swin                  & 87.30           & 39.41      & 2.19       \\
Swin V2               & 84.0           & 31.59      & 2.21       \\
TNT                   & 81.5           & 17.17      & 2.2        \\
TwinsPcPvt            & 83.2           & 30.11      & 2.18       \\
Visformer             & 78.6           & 26.03      & 2.21       \\
ViT                   & 85.17           & 11.25      & 2.61       \\
VOLO                  & 87.1           & 41.03      & 2.23       \\
XCiT                  & 83.4           & 40.69      & 2.2        \\
\hline
\end{tabular}
\end{table}

CaiT is a deeper transformer network for image classification that was created in the style of encoder/decoder architecture. Two improvements to the transformer architecture made by the author greatly increase the deep transformers' accuracy \cite{touvron2021going}. Using co-scale and Conv-attentional principles, Co-scale Conv-attentional image Transformers (CoaT) is a Transformer-based image classification \cite{xu2021co}. On ImageNet, CoaT achieves an accuracy of about 81.9\%. In Chen et al. CrossViT approach, image patches (i.e., tokens in a transformer) of various sizes are combined to create stronger image features using a dual-branch transformer \cite{chen2021crossvit}. CaiT, CoaT, and CrossViT all reports high ImageNet accuracy and fair performance on out-of-distribution datasets. Yet, CoaT achieves only half of top-1 accuracy on ImageNet-A, compared to CaiT and CrossViT. 

DeiT is a transformer without convolution that presents a teacher-student approach unique to transformers. It makes use of a distillation token to make sure that the student pays close attention to the teacher \cite{touvron2021training}. DeiT3 revisits the supervised training of ViTs and introduces a new straightforward data-augmentation method that uses just three augmentations, moving it closer to the self-supervised learning method \cite{touvron2022deit}. They achieve 81.8\% and 83.1\% accuracy on ImageNet respectively. DeiT3 reports the highest ImageNet-R and Stylized top-1 accuracy among all the models, with 53.63\% and 25.45\%  respectively.

As a design paradigm, EfficientFormer introduces a dimension-consistent pure transformer (without MobileNet blocks) and uses latency-driven slimming to produce many final models under EfficientFormer. The author demonstrates how EfficientFormer may shorten inference times and how well-designed transformers can achieve extremely low latency on mobile devices while achieving top performance \cite{li2022efficientformer}. While EfficientFormer reports 79.2\% accuracy on ImageNet, it does not perform well on out-of-distribution samples.

Hatamizadeh et al. proposed a global context vision transformer (GCViT), which improves parameter and compute usage for computer vision tasks that rely on global context self-attention modules combined with conventional local self-attention to accurately yet efficiently model both long and short-range spatial interactions as alternatives to complex operations like an attention mask or local windows shifting \cite{hatamizadeh2022global}. GCViT achieves the second highest ImageNet-A accuracy of 42.71\% and the third highest ImageNet-R and Sketch accuracy of 49.57\% and 36.29\% among Transformers.

The multi-axis attention paradigm, called MaxViT, is effective and scalable and has two components: blocked local and dilated global attention \cite{tu2022maxvit}. It obtains 86.5\% top-1 accuracy on ImageNet. MobileViT, a general-purpose vision transformer that is lightweight and mobile-friendly, obtains 78.4\% accuracy \cite{mehta2021mobilevit}. MobileViT's performance on ImageNet -A, R, Stylized, and Sketch are the lowest, yet it scores AUPR of 3.92 on ImageNet-O, the highest score among all the models. MaxViT outperforms MobileViT on robustness datasets.

\begin{table}[]
\caption{Performance of attention-based vision model on ImageNet-R, Stylized-ImageNet, and ImageNet-Sketch}
\label{tab:vit-2}
\begin{tabular}{cccc}
\hline
\textbf{Architecture} & \textbf{\begin{tabular}[c]{@{}c@{}}ImageNet\\ -R\end{tabular}} & \textbf{\begin{tabular}[c]{@{}c@{}}Stylized\\ -ImageNet\end{tabular}} & \textbf{\begin{tabular}[c]{@{}l@{}}ImageNet\\ -Sketch\end{tabular}} \\ \hline
BEiT            & 10.25    & 2.05        & 2.05  \\
CaiT            & 49.95    & 19.48       & 36.03 \\
CoaT            & 42.81    & 14.27       & 28.12 \\
CrossViT        & 47.10    & 18.51       & 33.49 \\
DeiT            & 45.36    & 17.99       & 32.35 \\
DeiT3           & 53.63    & 25.45       & 40.06 \\
EfficientFormer & 43.4     & 12.21       & 30.91 \\
GCViT           & 49.57    & 17.81       & 36.29 \\
MaxViT          & 47.8     & 17.75       & 35.19 \\
MobileViT       & 0.57     & 0.12        & 0.11  \\
MViTv2          & 51.89    & 19.16       & 38.11 \\
NesT            & 46.57    & 18.27       & 33.36     \\
PiT             & 44.37    & 16.4        & 32.70      \\
PVT V2          & 44.21    & 10.02       & 21.72      \\
Swin            & 48.64    & 18.36       & 34.37      \\
Swin V2         & 43.76    & 13.54       & 30.99      \\
TNT             & 39.23    & 12.56       & 21.03      \\
TwinsPcPvt      & 45.92    & 15.97       & 32.57      \\
Visformer       & 43.39    & 13.58       & 30.11      \\
ViT             & 38.42    & 14.37       & 17.36      \\
VOLO            & 48.53    & 17.74       & 36.11      \\
XCiT            & 46.51    & 15.28       & 34.53      \\ \cline{1-4}
\end{tabular}
\end{table}

A unified architecture for image and video classification called Multiscale Vision Transformer (MViT2) uses residual pooling connections and decomposed relative positional embeddings \cite{li2022mvitv2}. On both ImageNet and small datasets like CIFAR, Zhang et al. Nested Hierarchical Transformer (NesT) converges faster and needs less training data to achieve good generalization. NesT results in a robust decoder eight times faster when applying key concepts to image generation than earlier transformer-based generators \cite{zhang2022nested}. With the incorporation of a pooling layer, the pooling-based Vision Transformer (PiT) improves model capability, and generalization performance in comparison to ViT \cite{heo2021rethinking}. MViT2 obtains the highest ImageNet score with 88.8\% accuracy, and NesT and PiT also obtain more than 81\% accuracy. MViT2 reports the second highest ImageNet-R score with 51.89\% top-1 accuracy and fairly high scores on out-of-distribution classification. NesT and PiT also performs well on various datasets. 

With the addition of a linear complexity attention layer, overlapping patch embedding, and a convolutional feed-forward network, PVTV2 \cite{wang2022pvt} is an improved version of the original Pyramid Vision Transformer (PVTV1) \cite{wang2021pyramid}. Although it obtains 81\% ImageNet score, it decrease drastically on robustness datasets, only obtaining 4.05\% ImageNet-A accuracy.

Swin is a hierarchical transformer whose representation is computed via shifted windows. This method increases performance by restricting self-attention computation to non-overlapping local windows while allowing a cross-window connection \cite{liu2021swin}. SwinV2 propose a residual-post-norm method coupled with cosine attention to enhance training stability, a log-spaced continuous position bias method to successfully transfer models pre-trained using low-resolution images to downstream tasks with high-resolution input, and a self-supervised pretraining method, SimMIM, to lessen the requirement for large labeled image datasets \cite{liu2022swin}. They both reports 87.3\% and 84\% accuracy on ImageNet. Interestingly, Swin outperforms SwinV2 on all of the out-of-distribution datasets.

TNT (Transformer iN Transformer) is a novel neural architecture that encodes input data into powerful features using the attention mechanism \cite{han2021transformer}. TwinsPcPvt explores the design of spatial attention and shows that a well-developed yet simple spatial attention mechanism outperforms state-of-the-art techniques \cite{chu2021twins}. They achieve 81.5\% and 83.2\% top-1 accuracy respectively. Visformer, a Transformer-based visual recognition model, achieves 78.6\% top-1 accuracy on ImageNet \cite{chen2021visformer}.  TwinsPcPvt outperforms other two models on other datasets, and TNT does not perform well on ImageNet-A and Sketch.

Vision Transformer (ViT) proposed by Dosovitskiy et al., is a pure transformer that can be applied directly to sequences of picture patches and performs very well on image classification tasks \cite{dosovitskiy2020image, steiner2021train}. It achieves 85.17\% top-1 accuracy on ImageNet, but its performance on robustness datasets is not satisfactory. It scores only 17.36\% on ImageNet-Sketch, which is even lower than most of CNNs, and 11.25\% accuracy on ImageNet-A, one of the lowest score among Transformers.

Yuan et al. introduce novel viewpoint attention and present Vision Outlooker (VOLO), a basic and universal neural network architecture \cite{yuan2022volo}. XCiT is a transposed version of self-attention that operates across feature channels rather than tokens, where the interactions are based on the cross-covariance matrix between keys and queries \cite{ali2021xcit}. Both models achieve high scores on most of the datasets. VOLO obtains the second-highest ImageNet-A top-1 accuracy, along with the third-highest ImageNet-Sketch top-1 accuracy among Transformers. XCiT also reports the third highest ImageNet-A accuracy.

\subsection{Hybrid (combination of convolution and attention) Mechanism}

ConViT is a gated positional self-attention (GPSA) proposed by d'Ascoli et al., a type of positional self-attention that can be equipped with a "soft" convolutional inductive bias \cite{d2021convit}. EdgeNeXt is a hybrid architecture that combines the strengths of both the CNN and Transformer models. The author introduces the split depth-wise transpose attention (STDA) encoder, which divides input tensors into numerous channel groups and use depth-wise convolution and self attention across channel dimensions to implicitly extend the receptive field and encode multi-scale features \cite{maaz2022edgenext}. LeViT is a hybrid Vision Transformer in ConvNet's guise that optimizes the trade-off between accuracy and efficiency in a high-speed regime \cite{graham2021levit}. Each model obtains 81.3\%, 71.2\%, and 80\% accuracy on ImageNet. While EdgeNeXt achieves the lowest ImageNet accuracy among the three, it performs well on ImageNet-O, R, and Sketch. On the other hand, LeViT reports low scores on out-of-distribution samples despite its performance on ImageNet.

\begin{table}[]
\caption{Performance of a combination of CNN and attention-based vision models on ImageNet, ImageNet-A, and ImageNet-O}
\label{tab:comb-1}
\begin{tabular}{cccc}
\hline
\textbf{Architecture} & \textbf{ImageNet} & \textbf{ImageNet -A} & \textbf{\begin{tabular}[c]{@{}c@{}}ImageNet\\ -O\end{tabular}} \\ \hline

ConViT   & 81.30   & 29.69   & 2.21  \\
EdgeNeXt & 71.20   & 21.83   & 2.23  \\
LeViT    & 80.0   & 11.51   & 2.22  \\ \hline
\end{tabular}
\end{table}

\begin{table}[]
\caption{Performance of a combination of CNN and attention-based vision models on ImageNet-R, Stylized-ImageNet, and ImageNet-Sketch}
\label{tab:comb-2}
\begin{tabular}{cccc}
\hline
\textbf{Architecture} & \textbf{\begin{tabular}[c]{@{}c@{}}ImageNet\\ -R\end{tabular}} & \textbf{\begin{tabular}[c]{@{}c@{}}Stylized\\ -ImageNet\end{tabular}} & \textbf{\begin{tabular}[c]{@{}l@{}}ImageNet\\ -Sketch\end{tabular}} \\ \hline
ConViT   & 48.79  & 19.81      & 35.53    \\
EdgeNeXt & 51.04  & 16.26      & 37.13    \\
LeViT    & 40.28  & 15.59     & 26.39    \\ \hline
\end{tabular}
\end{table}

\subsection{Multi-layer Perceptron Mechanism}

MLP-Mixer, a multi-layer perceptron (MLP)-based architecture with two types of layers: one with MLPs applied individually to image patches (i.e. "mixing" per-location characteristics) and one with MLPs applied across patches (i.e. "mixing" spatial information) \cite{tolstikhin2021mlp}. MLP-Mixer obtains 76.44\% top-1 accuracy on ImageNet, but its performance decreases on other datasets. It only scores 5.79\% on ImageNet-A and 10.04\% on Style, lower than most of CNNs or Transformers.

\subsection{Complementary Search Mechanism}

MobileNetV3 is based on a complimentary search technique as well as a novel architecture design customized to mobile phone CPUs via a mix of hardware-aware network architecture search (NAS) augmented by the NetAdapt algorithm and then improved through novel architecture developments \cite{howard2019searching}. It obtains 75.77\% accuracy on ImageNet, but similarly to MLP-Mixer, it only achieves 4.59\% and 10.85\% on ImageNet-A and Style.

\subsection{Others Mechanism}

RegNet is a two-dimensional design space made of simple, regular networks \cite{radosavovic2020designing}. Sequencer is a unique and competitive architecture representing long-term dependencies using LSTMs rather than self-attention layers \cite{tatsunami2022sequencer}. 
Both RegNet and Sequencer obtain more than 80\% top-1 accuracy on ImageNet, and demonstrate high score on robustness datasets. Especially, Sequencer outperforms most of the convolution-based and attention-based models.

\begin{table}[]
\caption{Performance of a combination of other (not CNN or attention-based) models on ImageNet, ImageNet-A, and ImageNet-O}
\label{tab:other-1}
\begin{tabular}{cccc}
\hline
\textbf{Architecture} & \textbf{ImageNet} & \textbf{\begin{tabular}[c]{@{}c@{}}ImageNet\\ -A\end{tabular}} & \textbf{\begin{tabular}[c]{@{}c@{}}ImageNet\\ -O\end{tabular}} \\ \hline
MLPMixer  & 76.44   & 5.79   & 2.25  \\
MobileNetV3 & 75.77   & 4.59   & 2.2   \\
RegNet    & 80.25    & 26.48  & 2.17  \\
Sequencer & 84.60   & 35.36  & 2.18  \\ \hline
\end{tabular}
\end{table}

\begin{table}[]
\caption{Performance of a combination of other (not CNN or attention-based) models on ImageNet-R, Stylized-ImageNet, and ImageNet-Sketch}
\label{tab:other-2}
\begin{tabular}{cccc}
\hline
\textbf{Architecture} & \textbf{\begin{tabular}[c]{@{}c@{}}ImageNet\\ -R\end{tabular}} & \textbf{\begin{tabular}[c]{@{}c@{}}Stylized\\ -ImageNet\end{tabular}} & \textbf{\begin{tabular}[c]{@{}l@{}}ImageNet\\ -Sketch\end{tabular}} \\ \hline
MLPMixer  & 32.75  & 10.04   & 19.32  \\
MobileNetV3 & 35.11  & 10.85   & 22.95 \\
RegNet    & 44.74  & 11.88  & 32.93  \\
Sequencer & 48.56  & 16.74    & 35.87 \\ \hline
\end{tabular}
\end{table}

\section{Discussion and Conclusion}

Vision transformers are becoming state-of-the-art image recognition tasks, and researchers are investigating their robustness compared to CNNs. Recent studies indicated that transformers are more robust than CNN; however, our observations indicate that the analysis is not fair and not particularly deep because the available literature only considers one or two basic models for comparison from each family. However, we offered comprehensive robustness research comparing 58 computer vision models not limited to convolution or attention families. We discover that if CNN is given the same training scenario as the transformer, it is just as resilient. Among all the 58 models, DeiT3 performs best in terms of robustness due to its teacher-student approach. ConvNext, a newly developed convolution-based model, outperforms out-of-distribution data. Our research also reveals that the convolution and attention model performs similarly to transformers. Aside from convolution and attention-based models, the sequencer (sequence-based model) is robust. On out-of-distribution data, base versions of state-of-the-art CNN models like ResNet and VGG perform poorly. This work provides a thorough robustness analysis of almost each available state-of-the-art computer vision model based on almost all available robustness data. In the future, we hope to look at the behavior of neural networks when they classify data from distributions. Looking at the problem from a causal standpoint, mainly studying the spurious correlation between non-robust features, may be helpful in designing a robust computer vision model with generalization capabilities.







{\small
\bibliographystyle{ieee_fullname}
\bibliography{egbib}
}

\end{document}